\documentclass[ letterpaper, 10pt, conference]{ieeeconf}      

\IEEEoverridecommandlockouts                              

\usepackage{amsmath,amssymb} 
\usepackage{graphicx}
\usepackage{textcomp}
\usepackage{graphics} 
\usepackage{epsfig} 
\usepackage{subfigure}
\usepackage{capt-of}
\usepackage{amsmath} 
\usepackage{amssymb}  
\usepackage{epstopdf}
\usepackage{lipsum}
\usepackage{lipsum,amsmath,multicol}
\usepackage{float}
\usepackage[linesnumbered,ruled,vlined]{algorithm2e}
\usepackage{algorithm2e}

\SetCommentSty{mycommfont}

\SetKwInput{KwInput}{Input}        
\SetKwInput{KwReturn1}{Return}    
\SetKwInput{KwOutput}{Output} 
\SetKwInput{KwInitialization}{Initialization} 
\usepackage{wrapfig}
\usepackage{sidecap}

\usepackage{amsthm}

\usepackage{comment}
\usepackage{array}
\usepackage{glossaries}
\usepackage{breqn}
\makeglossaries
\newcolumntype{L}[1]{>{\raggedright\let\newline\\\arraybackslash\hspace{0pt}}m{#1}}
\newcolumntype{C}[1]{>{\centering\let\newline\\\arraybackslash\hspace{0pt}}m{#1}}
\newcolumntype{R}[1]{>{\raggedleft\let\newline\\\arraybackslash\hspace{0pt}}m{#1}}
\usepackage{sidecap}
\usepackage{url}
\usepackage[T1]{fontenc}
\usepackage{tabularx}
\usepackage{multicol, blindtext}
\usepackage{amsmath}
\usepackage{multirow}
\usepackage[table,xcdraw]{xcolor}
\usepackage{graphicx}
\usepackage[skip=2pt,font=footnotesize]{caption}

\usepackage{mathtools, bigstrut}
\usepackage{tabularx}
\usepackage{multicol, blindtext}
\usepackage{amsmath}
\usepackage{multirow}

\newtheorem{remak}{Remark}
\allowdisplaybreaks

\newcommand\blfootnote[1]{%
  \begingroup
  \renewcommand\thefootnote{}\footnote{#1}%
  \addtocounter{footnote}{-1}%
  \endgroup
}

\begin{document} 

\let\ACMmaketitle=\maketitle

\title{PRIEST: Projection Guided Sampling-Based Optimization For Autonomous Navigation}
\author{ Fatemeh Rastgar, Houman Masnavi, Basant Sharma, Alvo Aabloo, Jan Swevers, Arun Kumar Singh}
\maketitle

\blfootnote{Fatemeh Rastgar, Houman Masnavi, Basant Sharma, Alvo Aabloo, and Arun Kumar Singh are with the Institute of Technology, University of Tartu. Jan Swevers is with the Mechanical Engineering department at KU Leuven University and Flanders Make KU Leuven. This research has been financed by European Social Fund via ICT program measure and grants PSG753 from Estonian Research Council.}

\begin{abstract}
Efficient navigation in unknown and dynamic environments is crucial for expanding the application domain of mobile robots. The core challenge stems from the non-availability of a feasible global path for guiding optimization-based local planners. As a result, existing local planners often get trapped in poor local minima. In this paper, we present a novel optimizer that can explore multiple homotopies to plan high-quality trajectories over long horizons while still being fast enough for real-time applications. We build on the gradient-free paradigm by augmenting the trajectory sampling strategy with a projection optimization that guides the samples toward a feasible region. As a result, our approach can recover from the frequently encountered pathological cases wherein all the sampled trajectories lie in the high-cost region. Furthermore, we also show that our projection optimization has a highly parallelizable structure that can be easily accelerated over GPUs. We push the state-of-the-art in the following respects. Over the navigation stack of the Robot Operating System (ROS), we show an improvement of 7-13\% in success rate and up to two times in total travel time metric. On the same benchmarks and metrics, our approach achieves up to 44\% improvement over MPPI and its recent variants. On simple point-to-point navigation tasks, our optimizer is up to two times more reliable than SOTA gradient-based solvers, as well as sampling-based approaches such as the Cross-Entropy Method (CEM) and VPSTO. Codes: \url{https://github.com/fatemeh-rastgar/PRIEST}
\end{abstract}

\section{Introduction} \label{Introduction}
Smooth and collision-free navigation in unknown and dynamic environments is crucial for the deployment of mobile robots in places like hospitals, warehouses, airports, etc. In these human-habitable environments, the layout of the static obstacles can change over time. Moreover, human movement can create additional dynamic obstacles obstructing the robot's movements. As a result, the prior computed global plan invariably becomes infeasible during the robot's motion and can no longer guide the local planner toward safe state-space regions. One possible workaround is to make the local planners themselves capable of planning over long horizons while exploring different trajectory homotopies in real time. Our work is geared towards imparting such capabilities to mobile robots.

\begin{figure}
    \centering
    \includegraphics[scale=0.30
    ]{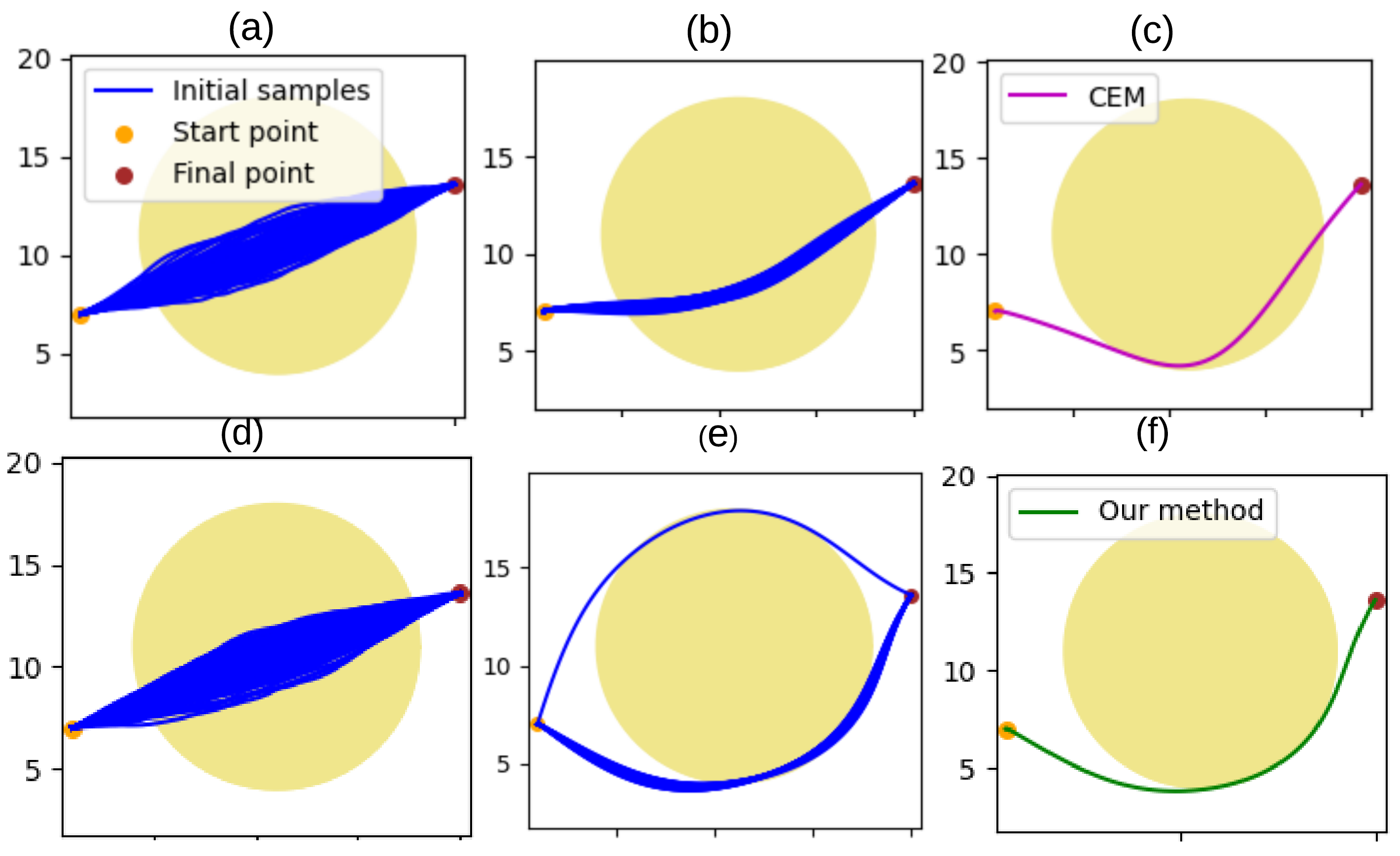}
    \vspace{-0.1cm}
    \caption{A comparison of CEM and our proposed approach. Fig.(a)-(c) shows how typical CEM (or any sampling-based optimizer) struggles when all the sampled initial trajectories lie in the high-cost/infeasible region. Our approach, PRIEST, embeds a projection optimizer within any standard sampling-based approach that pushes the samples toward feasible regions before evaluating their cost.}
 \vspace{-0.75cm}
    \label{fig_teaser}
\end{figure}

In this paper, we consider optimization-based local planners because of their ability to satisfy constraints and produce smooth motions. Moreover, this approach also allows us to encode some desired higher-level behaviors through appropriately designed cost functions. There are two broad classes of approaches to solving optimization problems encountered during trajectory planning. On one end of the spectrum, we have gradient-based approaches \cite{metz2003rockit, vanroye2023fatrop} that require the cost and constraint functions to be differentiable. Typically, these methods depend heavily on the user providing a good guess for the solution to initialize the optimization solver. However, finding good trajectory initializations is challenging in fast-changing environments. Some approaches, e.g., based on integer programming \cite{cplex}, can cope with poor initialization. But they are typically computationally too slow, especially in very cluttered environments with tens of obstacles.  

On the other end of the spectrum, we have planners based on sampling-based optimizers such as Cross-Entropy Method (CEM) \cite{bharadhwaj2020model} and Covariance Matrix Adaptation-Evolution Strategy (CMA-ES) \cite{jankowski2023vp}. These optimizers perform a random sampling of the state-space trajectories to obtain a locally optimal solution. Due to this exploration property, they can often come up with better solutions than purely gradient-based approaches \cite{petrovic2020cross}.  Moreover, sampling-based optimizers are easily parallelizable and thus can be accelerated over GPUs. However, one fundamental drawback is that these optimizers typically fail when all the sampled trajectories lie in the high-cost region (refer Fig.\ref{fig_teaser}(a-c)).

Our main motivation in this paper is to combine the benefits of both sampling-based and gradient-based approaches. Existing efforts in this direction are mostly restricted to using sampling-based optimizers for computing a good initialization trajectory, which is then subsequently fed to the gradient-based solver \cite{kim2022mppi}. Our experiments in Section \ref{comb_grad_sample} show that such approaches do not work reliably in difficult benchmarks since they still inherit the issues prevalent with both individual classes of approaches. Moreover, at a more fundamental level, the sampling optimizer is unaware of the capabilities of the downstream gradient-based solver or how it refines the initial guess.

Our main idea in this paper is to use gradient-based approaches to improve the inner working of sampling-based optimization. In particular, we formulate a projection optimization to guide the sampling process toward low-cost regions at each iteration. As a result, our approach can recover from the pathological cases where all sampled trajectories are infeasible, e.g., due to violation of collision constraints(see Fig.\ref{fig_teaser}(d-f)). Our core innovations and their benefits are summarized below.

\noindent \textbf{Algorithmic Contribution:} We present Projection Guided Sampling Based Optimization (PRIEST). The key building block is a novel optimizer that can take a set of trajectories and project each of them onto the feasible set. This allows us to guide sampled trajectories toward feasible regions before evaluating their cost and subsequent refinement of the sampling distribution. We show how our projection optimizer can be effectively parallelized and accelerated over GPUs by reformulating the underlying collision and kinematic constraints into polar form and using an Alternating Minimization (AM) approach to the resulting problem. Finally, we show how our projection optimizer naturally integrates with decentralized variants of sampling-based optimizers \cite{zhang2022simple}, wherein multiple sampling distributions are refined in parallel to improve the optimality of the solution. See Section \ref{rel_works} for a summary of contributions over authors' prior works.

\noindent \textbf{Improvement over the State-of-the-art (SOTA):} We show that PRIEST outperforms existing approaches in terms of success rate, time-to-reach the goal, and computation time, etc. In particular, we show at least 7\% improvement over the ROS Navigation stack in success~rate on the  BARN dataset~\cite{perille2020benchmarking}, while reducing the travel time by a factor of two. On the same benchmarks, our success rate is at least 35\% better than SOTA local sampling-based optimizers like MPPI \cite{williams2017model} and log-MPPI~\cite{mohamed2022autonomous}. Additionally, we consider a point-to-point navigation task and compare PRIEST with the SOTA gradient-based solvers, ROCKIT \cite{metz2003rockit}(a collection of optimizers like IPOPT, ACADO, etc) and FATROP~\cite{vanroye2023fatrop}, and sampling-based methods CEM and VPSTO~\cite{jankowski2023vp}. We show up to $2$x improvement in success rate over these baselines. Furthermore, we show that PRIEST respectively has 17\% and 23\% higher success rates than the ROS Navigation stack and other SOTA approaches in dynamic environments.  

\section{Mathematical Preliminaries}
\subsubsection*{Symbols and Notations}
Small case letters with regular and bold font represent scalars and vectors, respectively. Matrices have upper-case bold fonts. The variables $t$ and $T$ are time stamps and transpose, respectively. The number of planning steps, obstacles, decision variables, and samples are shown as $n_{p},n_{o},n_{v}$ and $N_{b}$. The left subscript $k$ denotes the trajectory optimizer's iteration. The rest of the symbols will be defined in the first place of use. 

\subsection{Problem Formulation}
\subsubsection{Differential Flatness} We leverage differential flatness to make our approach applicable to a large class of systems such as wheeled mobile robots, quadrotors, etc. Specifically, we assume $\hspace{-0.02cm}\textbf{u}\hspace{-0.12cm}=\hspace{-0.08cm}\boldsymbol{\Phi}(\hspace{-0.03cm}x^{(q)}\hspace{-0.02cm}(t),\hspace{-0.04cm}y^{(q)}\hspace{-0.04cm}(t),z^{(q)}\hspace{-0.02cm}(t)\hspace{-0.07cm})$: the control inputs can be obtained through some analytical mapping $\boldsymbol{\Phi}$ of $q^{th}$ level derivatives of the position-level trajectory. For example, for a quadrotor, the pitch, roll, yaw angles, and thrust can be analytically expressed in terms of axis-wise accelerations. 

\subsubsection{Trajectory Optimization}
We are interested in solving the following 3D trajectory optimization:

\vspace{-0.37cm}
\small
\begin{subequations}
\begin{align}
   \hspace{-2.5cm}\min_{x(t), y(t), z(t)} c_1(x^{(q)}(t), y^{(q)}(t), z^{(q)}(t)),  
\label{acc_cost}\\
   x^{(q)}(t), y^{(q)}(t), z^{(q)}(t)|_{t=t_{0}} = \textbf{b}_{0},\nonumber \\
   x^{(q)}(t), y^{(q)}(t), z^{(q)}(t)|_{t=t_{f}} = \textbf{b}_{f},\label{eq1_multiagent_1}\\
    \dot{x}^{2}(t) + \dot{y}^{2}(t) + \dot{z}^{2}(t) \leq v^{2}_{max}, \nonumber \\
    \ddot{x}^{2}(t) + \ddot{y}^{2}(t) + \ddot{z}^{2}(t) \leq a^{2}_{max}, \label{acc_constraint}\\
    \hspace{-0.4cm}
    s_{min} \leq (x(t), y(t), z(t)) \leq s_{max} \label{affine_ineq}  \\
    \hspace{-0.4cm}
    -\frac{\hspace{-0.07cm}(x(t)\hspace{-0.09cm}-
 \hspace{-0.06cm}x_{o, j}(t)\hspace{-0.01cm})^{2}}{a^2}\hspace{-0.09cm} - \hspace{-0.09cm}
 \frac{\hspace{-0.07cm}(y(t) \hspace{-0.075cm}-\hspace{-0.065cm}y_{o, j}(t)\hspace{-0.02cm})^{2}}{a^2} 
  \hspace{-0.09cm}-\hspace{-0.09cm}\frac{\hspace{-0.07cm}(z(t)\hspace{-0.09cm}- 
 \hspace{-0.08cm}z_{o, j}(t)\hspace{-0.01cm})^{2}}{b^2}\hspace{-0.06cm} \hspace{-0.06cm}
  + \hspace{-0.05cm}1\leq 0, \label{coll_multiagent}
\end{align}
\end{subequations}
\normalsize

\noindent where $(x(t),y(t),z(t))$ and $(x_{o, j}(t),y_{o, j}(t),z_{o, j}(t))$ respectively denote the robot and the $j^{th}$ obstacle position at time $t$. The function $c_1(.)$ is defined in terms of derivatives of the position-level trajectories and can encompass commonly used penalties on accelerations, velocities, curvature, etc. We can also leverage differential flatness to augment control costs in $c_1(.)$ as well. The affine inequalities \eqref{affine_ineq} model bounds on the robot workspace. The vectors $\textbf{b}_{0}$ and $\textbf{b}_{f}$ in \eqref{eq1_multiagent_1} represent the initial and final values of boundary condition on the $q^{th}$ derivative of the~position-level trajectory. In our formulation, $q\hspace{-0.10cm}=\hspace{-0.10cm}\{0,1,2\}$. Inequalities \eqref{acc_constraint} denotes the velocity and acceleration bounds with their respective maximum values being $v_{max}$ and $a_{max}$. In \eqref{coll_multiagent}, we enforce collision avoidance, assuming obstacles are modeled as axis-aligned ellipsoids with dimensions $(a, a, b)$.

\begin{remak}
    The cost functions $c_1(.)$ need not be convex, smooth or even have an analytical form in our approach.
\end{remak}

\subsubsection{Trajectory Parametrization and Finite Dimensional Representation }
To ensure smoothness in the trajectories, we parametrize the optimization variables $(x(t),y(t),z(t))$~as 

\vspace{-0.32cm}
\small
\begin{align}
    \begin{bmatrix}
    x(t_{1}) \\ \vdots \\ x(t_{n_{p}})
    \end{bmatrix}^{T} \hspace{-0.3cm} = \bold{P} \hspace{0.1cm}\bold{c}_{x} ,\hspace{-0.1cm}
     \begin{bmatrix}
    y(t_{1})\\ \vdots\\ y(t_{n_{p}})
    \end{bmatrix}^{T} \hspace{-0.2cm}= \bold{P} \bold{c}_{y}, \hspace{-0.1cm}
    \begin{bmatrix}
    z(t_{1}) \\ \vdots \\ z(t_{n_{p}})
    \end{bmatrix}^{T} \hspace{-0.3cm} = \bold{P} \hspace{0.1cm}\bold{c}_{z} \label{parametrized}
\end{align}
\normalsize

\noindent where $\bold{P}$ is a matrix created using polynomial basis functions that are dependent on time and $\bold{c}_{x}, \bold{c}_{y}, \bold{c}_{z}$ represent the coefficients of the polynomial. The expression remains applicable for derivatives by utilizing  $\dot{\bold{P}}$ and $\ddot{\bold{P}}$. 

By incorporating the parametrized optimization variables stated in \eqref{parametrized} and compact representation of variables, we can reframe the  optimization problem \eqref{acc_cost}-\eqref{coll_multiagent} as follows:

\vspace{-0.37cm}
\small
\begin{subequations}
\begin{align}   \min_{\hspace{0.1cm}\boldsymbol{\xi}} c_1(\boldsymbol{\xi})  \label{cost_modify} \\
    \textbf{A}\boldsymbol{\xi} = \textbf{b}_{eq}\label{eq_modify}\\
   \textbf{g}(\boldsymbol{\xi}) \leq \textbf{0} \label{reform_bound},
\end{align}
\end{subequations}
\normalsize

\noindent where $\hspace{-0.05cm}\boldsymbol{\xi}\hspace{-0.15cm}  = \hspace{-0.15cm}\begin{bmatrix}
\bold{c}_{x}^{T}&\hspace{-0.3cm}\bold{c}_{y}^{T}&\hspace{-0.3cm}
     \bold{c}_{z}^{T} 
  \end{bmatrix}^T\hspace{-0.20cm}$. With a slight abuse of notation, we have now used $c_1(.)$ to denote a cost function dependent on $\boldsymbol{\xi}$. The matrix $\textbf{A}$ is a block diagonal where each block on the main diagonal consists of $ \begin{bmatrix}
\bold{P}_{0}&\hspace{-0.22cm}\dot{\bold{P}}_{0}&\hspace{-0.22cm}\ddot{\bold{P}}_{0}&\hspace{-0.22cm} \bold{P}_{-1}
\end{bmatrix}$. The subscript $0$, $-1$ signify the first and last row of the respective matrices and pertain to the initial and final boundary constraints.
The vector $\textbf{b}_{eq}$ is simply the stack of $\textbf{b}_{0}\hspace{-0.04cm}$ and $\textbf{b}_{f}$. The function $\textbf{g}$ contains all the inequality constraints \eqref{acc_constraint}-\eqref{coll_multiagent}. 

\section{Main Algorithmic Results}
\begin{figure}[h]
    \centering
    \includegraphics[scale=0.47
    ]{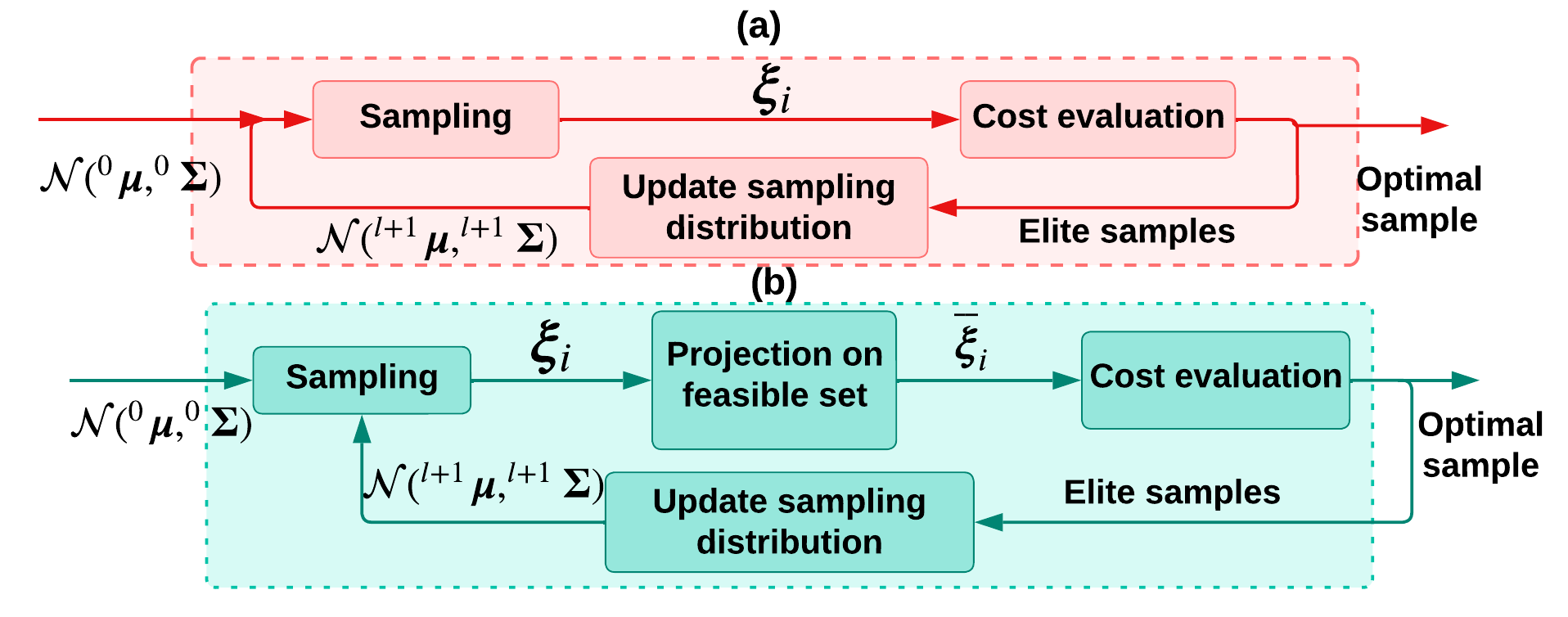}
    \caption{\hspace{-0.01cm}Comparison between a sampling-based optimizer~(a) and PRIEST~(b)}
    \label{fig_overview}
    \vspace{-0.4cm}
\end{figure}

This section presents our main algorithmic contributions. An overview of our approach is shown in Fig.\ref{fig_overview}. The main differentiating factor from existing baselines lies in the insertion of the projection optimizer between the sampling and cost evaluation block. The projection optimizer aids in constraint handling by pushing the sampled trajectories toward feasible regions. In this sense, our approach combines the benefits of both a gradient-free approach and those based on differentiable cost/constraint functions. As shown in Appendix \ref{append}, our projection block, in particular, leverages tools from convex optimization.

We next present our main building block: the projection optimizer, followed by its integration into a sampling-based optimizer.

\vspace{-0.3cm}

\subsection{Projection Optimization}
Consider the following optimization problem

\vspace{-0.4cm}
\small
\begin{align}
     \min_{\overline{\boldsymbol{\xi}}_{i}} \frac{1}{2}\Vert \overline{\boldsymbol{\xi}}_i-\boldsymbol{\xi}_{i}\Vert_2^2, ~ i=1,2,...,N_{b} 
    \label{projection_cost}\\
    \textbf{A}\overline{\boldsymbol{\xi}}_{i} = \textbf{b}_{eq}, \qquad \textbf{g}(\overline{\boldsymbol{\xi}}_{i}) \leq  \textbf{0} \label{projection_const}
\end{align}
\normalsize

\noindent The cost function \eqref{projection_cost} aims to minimally modify the $i^{th}$ sampled trajectory $\boldsymbol{\xi}_i$ to $\overline{\boldsymbol{\xi}}_{i}$ in order to satisfy the equality and inequality constraints. In Appendix \ref{append}, we show that for a certain class of constraint functions $\textbf{g}$ formed with quadratic and affine constraints, optimization \eqref{projection_cost}-\eqref{projection_const} can be reduced to the fixed-point iteration of the following form.

\vspace{-0.4cm}
\small
\begin{subequations}
    \begin{align}
    {^{k+1}}\textbf{e}_i, {^{k+1}}\boldsymbol{\lambda}_i = \textbf{h} ({^k} \overline{\boldsymbol{\xi}}_{i}, {^k}\boldsymbol{\lambda}_i ) \label{fixed_point_1}\\
\hspace{-0.2cm}{^{k+1}}\overline{\boldsymbol{\xi}}_{i} \hspace{-0.1cm}=\hspace{-0.1cm}\arg\min_{\overline{\boldsymbol{\xi}}_{i}} \frac{1}{2}\Vert \overline{\boldsymbol{\xi}}_{i}-\boldsymbol{\xi}_i\Vert_2^2 +\frac{\rho}{2} \left\Vert \textbf{F}\overline{\boldsymbol{\xi}}_{i} -{^{k+1}}\textbf{e}_{i} \right\Vert_2^2\hspace{-0.2cm}-\hspace{-0.1cm}{^{k+1}}\boldsymbol{\lambda}_i^T\overline{\boldsymbol{\xi}}_{i}, \nonumber \\ \qquad \textbf{A}\overline{\boldsymbol{\xi}}_{i} = \textbf{b}_{eq} \label{fixed_point_2}
\end{align}
\end{subequations}
\normalsize

\noindent In \eqref{fixed_point_1}-\eqref{fixed_point_2},  $\textbf{F}$ represents a constant matrix and $\textbf{h}$ is some closed-form analytical function. The vector ${^{k+1}}\boldsymbol{\lambda}_i$ is the Lagrange multiplier at iteration $k+1$ of the projection~optimization. We derive these entities in  Appendix \ref{append}. The main computational burden of projection optimization stems from solving the QP \eqref{fixed_point_1}. However, since there are no inequality constraints in \eqref{fixed_point_2}, the QP essentially boils down to an affine transformation of the following form:

\vspace{-0.3cm}
\small
\begin{subequations}
    \begin{align}
({^{k+1}}\overline{\boldsymbol{\xi}}_{i}, {^{k+1}}\boldsymbol{\nu}_i) = \textbf{M}\boldsymbol{\eta}({^k} \overline{\boldsymbol{\xi}}_{i}),
    \label{affine_trans} \\
    \textbf{M} \hspace{-0.1cm}= \hspace{-0.1cm}\begin{bmatrix}
        \textbf{I}+\rho\textbf{F}^T\textbf{F} & \textbf{A}^{T} \\ 
        \textbf{A} & \textbf{0}
    \end{bmatrix}^{-1}\hspace{-0.2cm}, \boldsymbol{\eta} = \begin{bmatrix}
        -\rho\textbf{F}^T {^{k+1}}\textbf{e}_i+{^{k+1}}\boldsymbol{\lambda}_i+\boldsymbol{\xi}_i\\
        \textbf{b}_{eq}
    \end{bmatrix} \label{matrix_vec}
\end{align}
\end{subequations}
\normalsize

\noindent \textbf{GPU Accelerated Batch Operation:} Our approach requires projecting several sampled trajectories onto the feasible set (see Fig.\ref{fig_overview}). However, this can be computationally prohibitive if done sequentially. Fortunately, our projection optimizer has certain structures that allow for batch/parallelized operation. To understand this further, note that the matrix $\textbf{M}$ in \eqref{affine_trans} is independent of the input trajectory sample $\boldsymbol{\xi}_{i}$. In fact, $\textbf{M}$ remains the same, irrespective of the trajectory sample that we need to project to the feasible set. This allows us to express the solution \eqref{affine_trans}, $\forall \boldsymbol{\xi}_{i}, i = 1,2,\dots,N_b$ as one large matrix-vector product, that can be trivially parallelized over GPUs. Similarly, acceleration can be done for the function $\textbf{h}$ that consists of just element-wise products and sums.

\begin{figure}[!b]
\vspace{-0.73cm}
    \centering
        \includegraphics[scale=0.32]{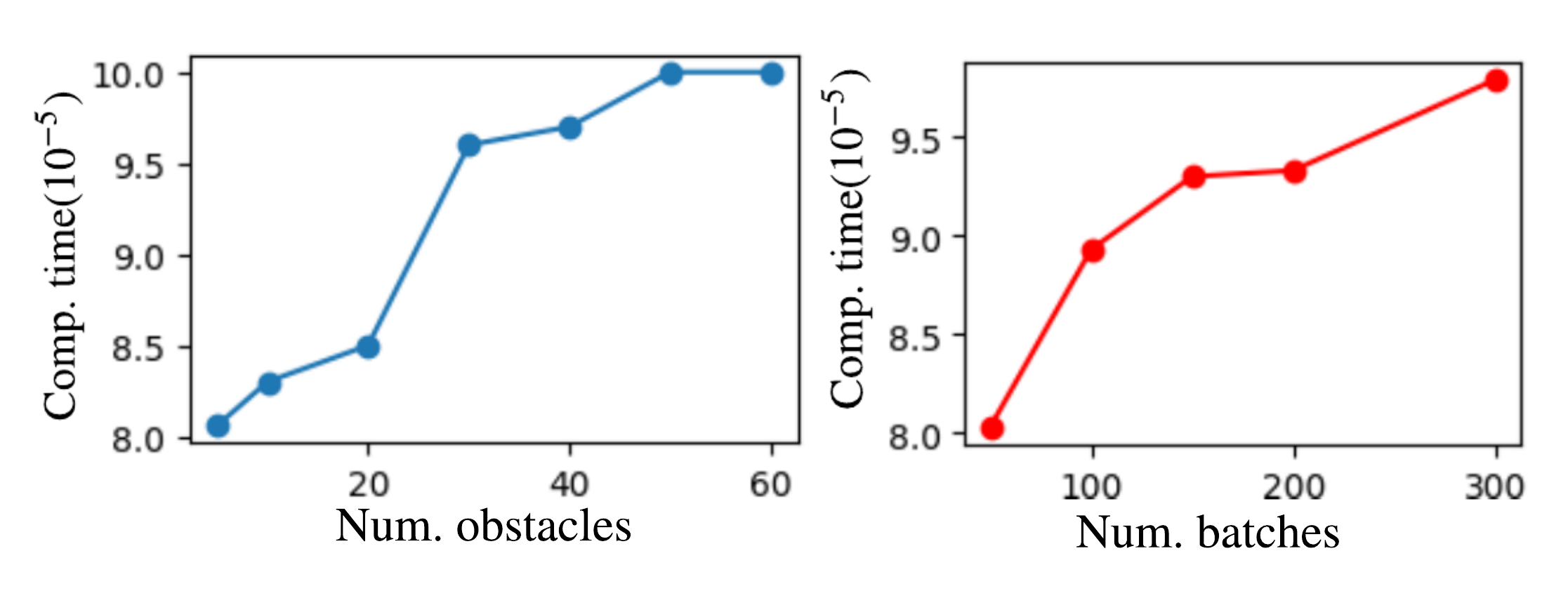}
    \caption{Scalability of per-iteration computation time of our projection optimizer with respect to number of obstacles and batch size}
    \label{comp_time}
    \vspace{-0.7cm}
\end{figure}

\noindent \textbf{Scalability:} The matrix $\textbf{M}$ in \eqref{affine_trans} needs to be computed only once as $\textbf{A}, \textbf{F}$ do not change across the projection iteration. The matrix $\boldsymbol{\eta}$ on the other hand, is recomputed at every iteration and its computation cost is primarily dominated by $\textbf{F}^{T}~{^{k+1}}\textbf{e}_{i}$. The number of rows in $\textbf{F}$ and $\textbf{e}$ increases linearly with planning horizon, number of obstacles or batch size (see Remark \ref{matrix_scaling}). This feature, coupled with GPU acceleration, provides excellent scalability to our approach for long-horizon planning in highly cluttered~environments. Fig.\ref{comp_time} shows how the average per-iteration time of the projection optimizer scales with the number of obstacles and batch size. Both graphs reflect the typical scaling pattern of matrix multiplication on GPUs \cite{fatahalian2004understanding}. 

\subsection{Projection Guided Sampling-Based Optimizer}
\subsubsection{Algorithm Description}Algorithm \eqref{algo1} presents the other core contribution of this paper. It starts by generating $N_{b}$ samples of polynomial coefficients $\boldsymbol{\xi}$ from a Gaussian distribution with $\mathcal{N}({^l}\boldsymbol{\mu},{^l}\boldsymbol{\Sigma})$ at iteration $l=0$ (line 3). The sampled $\boldsymbol{\xi}_i$ is then projected onto the feasible set (lines~4-5). Due to real-time constraints, it may not be possible to run the projection optimization till convergence. Thus, in line 6, we compute the constraint residuals $r(\overline{\boldsymbol{\xi}}_{i})$ associated with each sample. In line 7, we select the top $N_{proj}$ samples with the least constraint residuals and append them to the list $ConstraintEliiteSet$. In lines 8-9, we construct an appended cost, $c_{aug}$, by appending the residual to the primary cost function. We evaluate $c_{aug}$ on the $ConstraintEliiteSet$ samples.
In line 11, we once again select the top $N_{ellite}$ samples with the lowest $c_{aug}$ and append them to the list $ElliteSet$. Finally, in line 12, we update the distribution based on the samples of the $ElliteSet$ and the associated $c_{aug}$ values. The final output of the optimizer is the sample from the $ElliteSet$ with the lowest $c_{aug}$.
\begin{algorithm}[t]
\DontPrintSemicolon
\SetAlgoLined
\SetNoFillComment
\caption{Projection Guided Sampling-Based Optimization (PRIEST) }\label{algo1}
\KwInput{Initial states}
\KwInitialization{Initiate $^{l}\boldsymbol{\mu}$ and $^{l}\boldsymbol{\Sigma} $ at $i=0$}
\For{$l \leq N$}{
    Initialize $CostList = []$
    
    Draw $N_{b}$ samples $\boldsymbol{\xi}_{1},...,\boldsymbol{\xi}_{N_{b}}$ from $\mathcal{N}(^{l}\boldsymbol{\mu},^{l}\boldsymbol{\Sigma})$\tcp*[l]{\hspace{-0.2cm}Generating samples from a Gaussian distribution}
   
    Solve the inner convex optimizer to obtain $\overline{\boldsymbol{\xi}}_{i}$:
    \tcp*[l]{Pushing the samples into feasible trajectory space}

    \small
    \vspace{-1cm}
    \begin{subequations}
        \begin{align}
            \min_{\overline{\boldsymbol{{\xi}}}_{i}} \frac{1}{2} \left\Vert \overline{\boldsymbol{\xi}}_{i} - \boldsymbol{\xi}_{i}  \right\Vert ^{2}_{2},  \label{proj_cost} \\
        \bold{A}\overline{\boldsymbol{\xi}}_{i}= \bold{b}_{eq}\label{eq_modify_p} \\
            \textbf{g}(\overline{\boldsymbol{\xi}}_{i}) \leq \textbf{0} \label{reform_bound_p},
        \end{align}
    \end{subequations}
    \normalsize

    
    Compute the residuals set $r(\overline{\boldsymbol{\xi}}_{i})$ \tcp*[l]{Computing each sample residuals}
   
    $ConstraintElliteSet \leftarrow$ Select $N_{proj}$ samples from $r(\overline{\boldsymbol{\xi}}_{i})$ with the lowest values. \tcp*[l]{Select top samples with the lowest residuals}
    
    Evaluate the cost
    
    $c_{aug} \leftarrow  c_1(\overline{\boldsymbol{\xi}}_{i})+ r(\overline{\boldsymbol{\xi}}_{i})$
    \tcp*[l]{Cost evaluation}
    Append $cost$ to the $CostList$
    
    $ElliteSet \leftarrow$ Select $N_{ellite}$ top samples with the lowest cost obtained from the $CostList$.
    
    Update the new mean and covariance, $^{l+1}\boldsymbol{\mu}$~and~$^{l+1}\boldsymbol{\Sigma}$, using \eqref{update_mean}-\eqref{update_covariance}
    \tcp*[l]{Update mean and covariance}
}
\textbf{Return} $\overline{\boldsymbol{\xi}}_{i}$ corresponding to the lowest cost in $EliteSet$
\end{algorithm}

\begin{figure}
    \vspace{-0.5cm}
    \hspace{-0.2cm}
    \includegraphics[scale=0.47
    ]{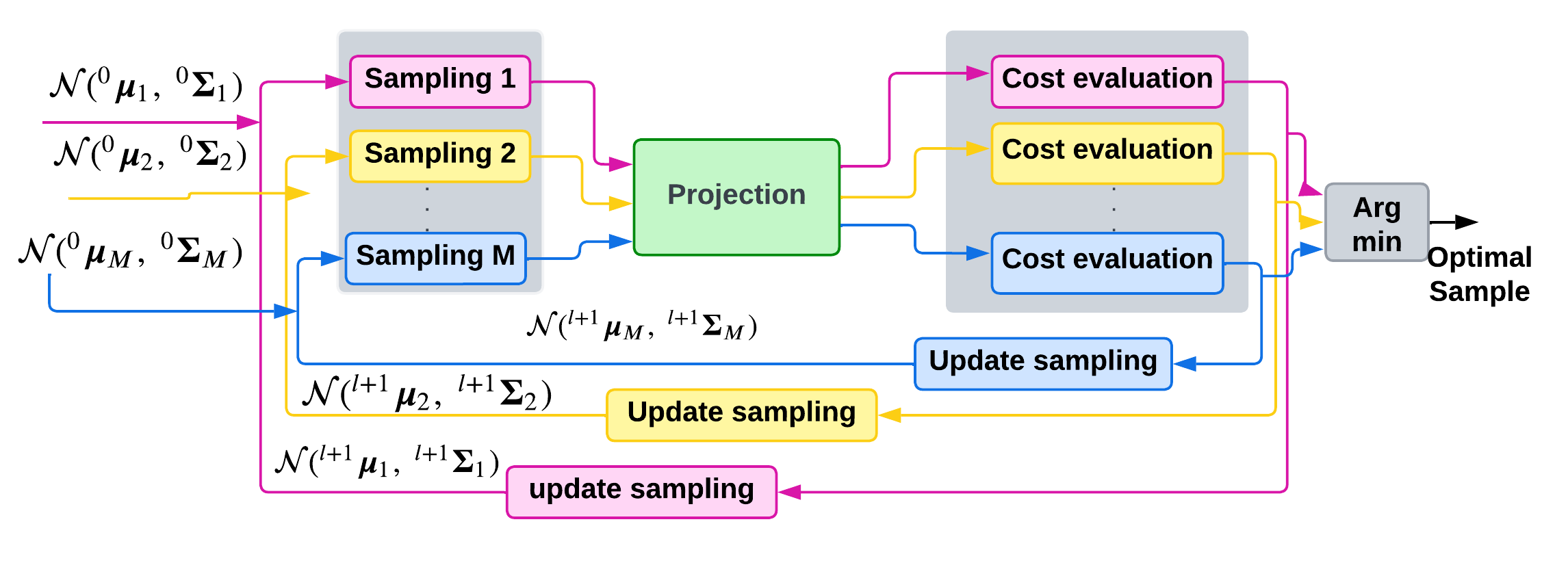}
    \caption{Decenteralized variant of PRIEST inspired by \cite{zhang2022simple}, wherein we instantiate and maintain different $M$ Gaussian distributions in parallel to counter poor local minima. An important thing to note is that our projection optimizer naturally fits into the decentralized structure.  }
    \label{fig_decentralized}
    \vspace{-0.8cm}
\end{figure}

\subsubsection{Updating the Sampling Distribution} There are several ways to perform the distribution update in line $13$ of Alg. \ref{algo1}. We adopt the following MPPI-like updated rule.






\vspace{-0.3cm}
 \small
    \begin{subequations}
        \begin{align}
         ^{l+1}\boldsymbol{\mu} = (1-\sigma)\hspace{0.1cm} ^{l}\boldsymbol{\mu} +\hspace{-0.05cm} \sigma (\frac{1}{\sum\limits_{m \in C} c_{m}})
         \sum\limits_{m\in C}\overline{\boldsymbol{\xi}}_{m} c_{m}, \label{update_mean} \\
        \hspace{-0.27cm}^{l+1} \boldsymbol{\Sigma}\hspace{-0.1cm} = \hspace{-0.1cm}(1\hspace{-0.1cm}-\hspace{-0.1cm}\sigma)  \hspace{0.05cm}^{l}\boldsymbol{\Sigma} \hspace{-0.1cm}+ \hspace{-0.1cm} \sigma \frac{\sum\limits_{m \in C} c_{m}(\overline{\boldsymbol{\xi}}_{m}-^{~l+1}\boldsymbol{\mu})(\overline{\boldsymbol{\xi}}_{m}-^{~l+1}\boldsymbol{\mu})^{T}}{\sum\limits_{m \in C} c_{m}}, 
         \label{update_covariance} \\
          c_{m} = \exp\big({\gamma^{-1}(c_{aug}(\overline{\boldsymbol{\xi}}_{m}) - \delta )}\big),
        \end{align}
    \end{subequations}
    \normalsize

\noindent where the scalar constant $\sigma$ is the so-called learning rate. The set $C$ consists of the top $N_{elite}$ selected trajectories (line 11). The constant $\gamma$ specifies the sensitivity of the exponentiated cost function $c_{aug}(\overline{\boldsymbol{\xi}}_{m})$ for top selected trajectories. $\delta = \min c_{aug}(^{l}\overline{\boldsymbol{\xi}}_{m})$ is defined to prevent numerical instability. 

\begin{remak}
    Alg.\ref{algo1} is agnostic to the distribution update rule. For example, \eqref{update_covariance} can be replaced with CMA-ES style update and our initial experiments in this regard have shown good results.
\end{remak}



\subsection{Decenteralized PRIEST (D-PRIEST)}
\noindent In this subsection, we build upon \cite{zhang2022simple} and propose a decentralized variant of our projection-guided sampling-based optimizer. As shown in Fig.\ref{fig_decentralized}, in the decentralized variant, we instantiate several optimizers in parallel and choose the lowest cost-optimal solution among these. As a result, such variants are shown to be more capable of escaping poor local minima. Our key objective in this section is to show that our projection optimizer naturally integrates into decentralized optimizers built along the lines of \cite{zhang2022simple}.

Fig.\ref{fig_decentralized} shows our proposed approach. We initialize $M$ different Gaussian distributions ${^l}\boldsymbol{\mu}_j, {^l}\boldsymbol{\Sigma}_j$ at $l=0$. We~sample $\frac{N_b}{M}$ samples of $\boldsymbol{\xi}$ from each of these distributions. The sampled $\boldsymbol{\xi}_{ij}$ ($i^{th}$ sample from $j^{th}$ Gaussian ) are then stacked row-wise to form a matrix. Importantly, such a construction allows us to easily track which row of the matrix corresponds to samples from which of the $M$ Gaussian distributions. The projection optimizer then simultaneously guides all the samples toward feasible regions. The output from the projection is then separated back into $M$ sets on which the cost functions are evaluated in parallel. We then update the sampling distribution in parallel based on the cost values. Finally, after $l$ iterations, the optimal trajectory from each optimizer instance is compared and the one with the lowest cost is selected as the optimal solution.

\section{Connections to Existing Works} \label{rel_works}
\noindent \textbf{Connection to CEM-GD:} Alternate approaches of combining sampling and gradient-based approach were presented recently in \cite{bharadhwaj2020model, huang2021cem}. In these two cited works, the projection at line 5 of Alg.\ref{algo1} is replaced with a gradient step of the form $\overline{\boldsymbol{\xi}}_i= \boldsymbol{\xi}_i-\sigma\nabla_{\boldsymbol{\xi}}c_1$, for some learning-rate $\sigma$. Our approach PRIEST improves \cite{bharadhwaj2020model, huang2021cem} in two main aspects. First, it can be applied to problems with non-smooth and non-analytical cost functions. Second, the gradient-descent-based can be computationally slow as it relies on taking small steps toward optimal solutions. In contrast, the projection optimizer in Alg.\ref{algo1} leverages convex optimization, specifically quadratic programming to ensure faster convergence.

\noindent \textbf{Exploration Strategy:} Sampling-based optimizers like MPPI \cite{williams2017model} and its variants \cite{mohamed2022autonomous,kim2022smooth} explore by injecting random perturbation into the control inputs to obtain a trajectory distribution. PRIEST, on the other hand, injects perturbation in the parameter space (polynomial coefficients), leading to one core advantage. We can inject larger perturbations into the parameter space that helps in better exploration over longer horizons (see Fig.\ref{qual_homo}(a)-(b)). Moreover, the projection optimizer ensures that the trajectory distribution satisfies boundary constraints and is pushed toward the feasible region. In contrast, increasing the covariance of control perturbation has been shown to make MPPI diverge~\cite{mohamed2022autonomous}.

\noindent \textbf{Improvement over Author's Prior Work \cite{masnavi2022visibility}}
PRIEST builds on our prior work \cite{masnavi2022visibility} that used projection-augmented sampling for visibility-aware navigation. Our current work targets a much broader scope of navigation problems with potentially non-smooth costs. On the algorithmic side, we extended the projection optimizer of \cite{masnavi2022visibility} to 3D (see Appendix \ref{append}) and improved the distribution update rule to account for actual cost values. Moreover, the decentralized variant of PRIEST is also a major contribution over \cite{masnavi2022visibility}. On the experimental side, we present a more elaborate benchmarking with several existing methods on both open-source as well as custom data sets.

\section{Validation and Benchmarking} \label{val}
\subsection{Implementation Details} 
\noindent We developed Alg.\ref{algo1}, PRIEST, in Python using JAX \cite{jax} library as our GPU-accelerated algebra backend. The simulation framework for our experiments was built on top of the ROS \cite{quigley2009ros} and utilized the Gazebo physics simulator. All benchmarks were executed on a Legion7 Lenovo laptop equipped with an Intel Core i7 processor and an Nvidia RTX 2070 GPU. Additionally, we used the open3d library to downsample PointCloud data \cite{zhou2018open3d}. For all the benchmarks, we chose $N =13,N_{b}= 110$, $N_{proj}=80$ and $N_{elite}=20$.

We develop a Model Predictive Control (MPC) pipeline on top of our optimizer. For each MPC iteration, 
we employ LIDAR scanning to infer obstacle locations. More specifically, we take each LIDAR point as an obstacle and employ voxel-downsampling through the open3d library to keep the number of obstacles tractable.

\subsubsection{{Baselines and Benchmarks}} 
We compared our approach with different baselines in three sets of benchmarks.

\noindent\textbf{Comparison on BARN Dataset \cite{perille2020benchmarking}:} This benchmark requires a mobile robot to iterative plan in a receding horizon fashion to navigate an obstacle field. We used the BARN dataset that contains 300 environments with varied levels of complexity, specifically designed to create local-minima traps for the robot. In this benchmark, we evaluate our approach against DWA \cite{fox1997dynamic}, TEB \cite{rosmann2017kinodynamic} implemented in the ROS navigation stack and MPPI \cite{williams2017model}, and log-MPPI \cite{mohamed2022autonomous}. The first two baselines are still considered the workhorse~of robot navigation in both industry and academia while the latter two are the SOTA gradient-free approaches for planning. No prior map was made available for any of the approaches. As a result, all approaches had to rely on their on-the-fly planning with a local cost map (or point cloud for our approach) for navigation. TEB and DWA used a combination of graph-search and optimization while MPPI and log-MPPI are purely optimization-based approaches. We used a holonomic robot modeled as a double-integrator system for the comparisons. Consequently, the cost function~($c_1$) used in our approach is given by a combination of penalty on the magnitude of the acceleration, curvature, and path-following error. The first two terms are smooth and differentiable and can be described in terms of axis-wise accelerations and velocities. The last term does not have an analytical form as it required computing the projection of a sampled trajectory way-point onto the straight-line path to the goal. 



\noindent \textbf{Point to Point Navigation with Differentiable Cost:}
In this benchmark, we considered the task of generating a single trajectory between a start and a goal location. The cost function  $c_1$ consisted of smoothness terms penalizing the norm of the acceleration. For comparison, we considered SOTA gradient-based optimizers ROCKIT~\cite{metz2003rockit} and FATROP \cite{vanroye2023fatrop} and sampling-based optimizers CEM, and VPSTO \cite{jankowski2023vp}. We designed various cluttered environments wherein obstacles are placed randomly in 2D and 3D spaces (see Fig.\ref{traj_plot}). For the 2D comparisons, our experiments involved 117 trials conducted in a cluttered environment featuring 50 randomly placed obstacles, each with a radius of $0.4$ m. For the 3D comparisons, we conducted $100$ trials in a cluttered room with dimensions of $7\times7\times3$ units and included $25$ randomly positioned obstacles, each with a radius of $0.68$ m.

\noindent \textbf{Comparison in a Dynamic Environment:} We also compare our approach PRIEST with CEM, log-MPPI, MPPI, TEB, and DWA in dynamic environments. The cost function used was the same as that used for the BARN dataset. In this benchmark, we introduced ten obstacles, each with a velocity of $0.1$ m/s, moving in the opposite direction of the robot. These obstacles have a radius of $0.3$ m. We run simulations over 30 different configurations. The start and final points remain constant for all the configurations, while the obstacle positions are varied for each configuration. The maximum velocity for the robot was fixed at $1$ m/s. 

\subsubsection{{Metrics}} 
We utilize the following metrics for benchmarking against other baselines:

\noindent \textbf{Success Rate}: A run is considered successful when the robot approaches the final point within a 0.5m radius without any collision. The success rate is calculated as the ratio of the total number of successful runs to the overall number of runs. 

\noindent \textbf{Travel Time}: This refers to the duration it takes for the robot to reach the vicinity of the goal point.

\noindent \textbf{Computation Time}: This metric quantifies the time required to calculate a solution trajectory.

\subsection{Qualitative Results}

\subsubsection{A Simple Benchmark}
\hspace{-0.08cm}In Fig.\eqref{fig_teaser}, our objective is to contrast the behavior of CEM(a-c) and our approach PRIEST(d-e) in a scenario wherein all the initial sampled trajectories are placed within a high-cost/infeasible region. To show this, we construct an environment with a large obstacle with a radius of $7$m. The task is to obtain a collision-free trajectory from $(1,7)$ to~$(20,13)$ while respecting the maximum velocity and acceleration limits of $2.8$ and $3.3$, respectively. As observed, our optimizer effectively pushes the sample trajectories out of the infeasible region. In contrast, the CEM samples persistently remain within the infeasible region. This behavior arises when the CEM variance fails and cannot navigate samples effectively out of the infeasible region.

\subsubsection{Receding Horizon Planning on Barn Dataset} 
\begin{figure}[!h]
    \centering
    \includegraphics[scale=0.50]{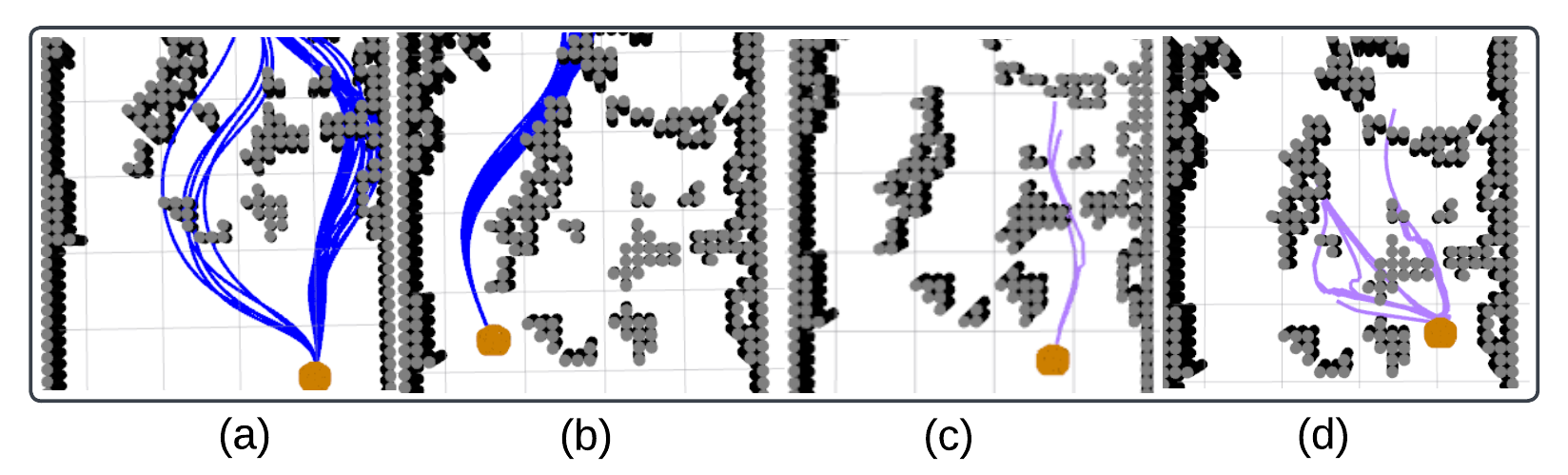}
    \caption{Qualitative result of MPC built on PRIEST~(a-b) and TEB (c-d).~Blue and purple trajectories show top samples in the PRIEST and TEB planner. }
    \label{qual_homo}
    \vspace{-0.4cm}
\end{figure}

In Fig. \eqref{qual_homo}, we show a qualitative comparison between TEB and our approach PRIEST within one of the BARN environments. Both methods can search over multiple homotopies but differ in their process. While TEB uses graph search, PRIEST relies on the stochasticity of the sampling process guided through the projection optimizer. As a result, the latter can search over a longer horizon and wider state space. It is worth pointing out that increasing the planning horizon of TEB dramatically increases the computation time and thus degrades the overall navigation performance instead of improving it. We present the quantitative comparison with TEB and other baselines in the next subsection.

\vspace{-0.1cm}


\subsubsection{Point-to-Point Navigation Benchmark} 
\begin{figure}[h]
    \centering
        \includegraphics[scale=0.42]{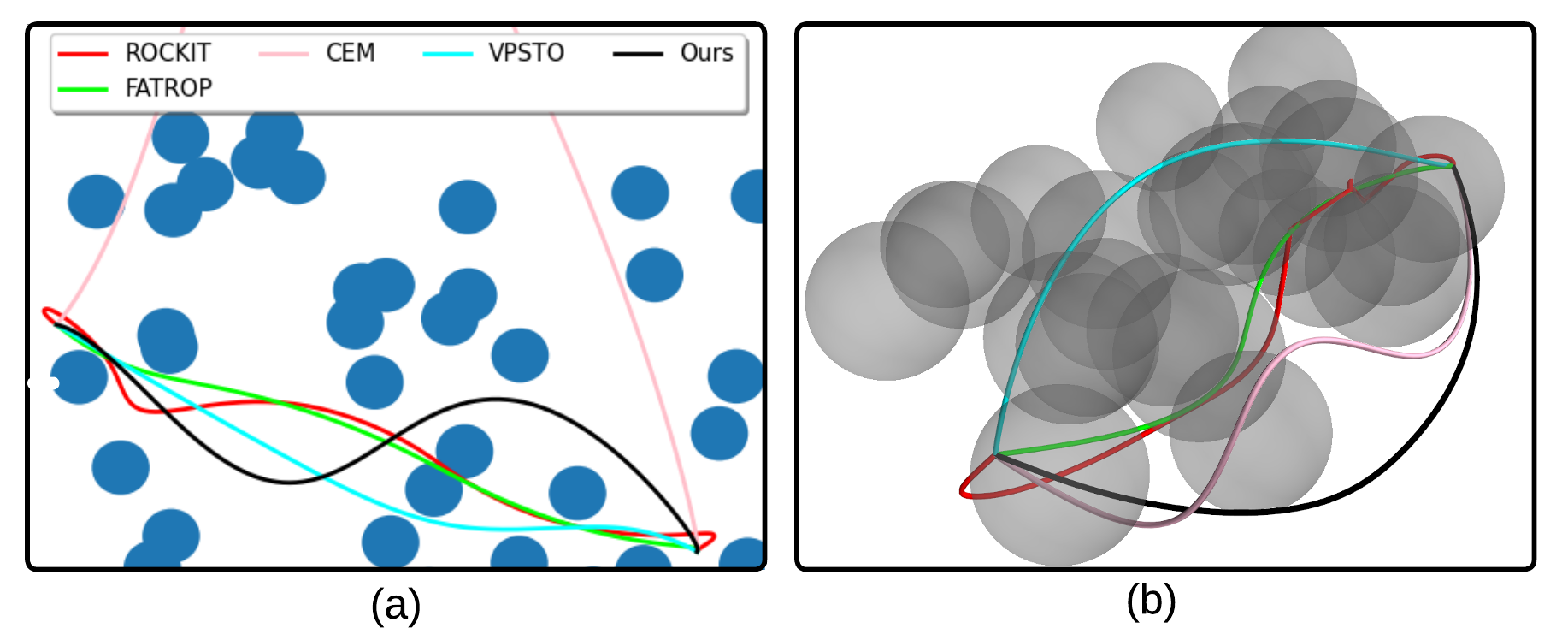}
    \caption{Qualitative result of point-to-point navigation. Trajectories for different approaches are shown in the same color for both 2D and 3D. }
    \label{traj_plot}
    \vspace{-0.5cm}
\end{figure}

\hspace{-0.12cm}Fig.\ref{traj_plot} shows trajectories generated by PRIEST alongside those generated by gradient-based optimizers, ROCKIT, FATROP, and sampling-based optimizers CEM, VPSTO. For the particular 2D example shown, both PRIEST and VPSTO successfully generated collision-free trajectories, while the other baselines failed. For the shown 3D environment, only PRIEST and CEM achieved collision-free trajectories. PRIEST trajectories were also smoother than other methods. In the next subsection, we present the quantitative statistical trends for all the baselines across different randomly generated environments.

\begin{figure}[h]
\vspace{-0.3cm}
\centering
    \includegraphics[scale=0.34]{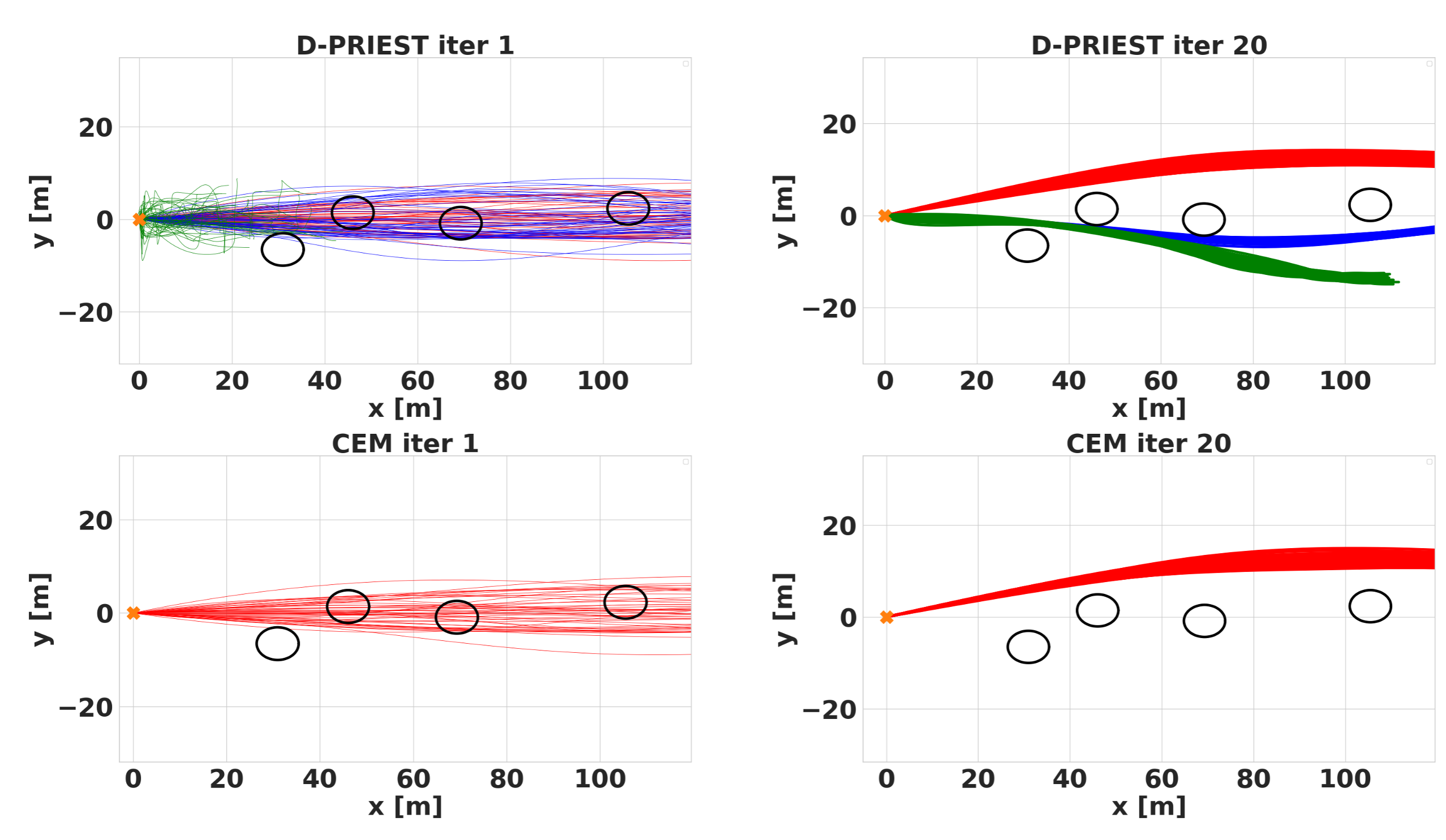}
    \caption{Comparison of D-PRIEST with baseline CEM. As is shown, the former updates multiple parallel distributions, resulting in multi-modal optimal trajectory distribution upon convergence. D-PRIEST maintained three different distributions (shown in green, red, and blue) and thus could obtain multi-modal behavior. In contrast, CEM which only has a single distribution, provides a limited set of options for collision-free trajectories.}
    \label{com_plot}
    \vspace{-0.4cm}
\end{figure}

\subsubsection{Decenteralized Variant} Fig.\ref{com_plot} shows the application of D-PRIEST for the trajectory planning of a car-like vehicle. The cost function $c_1$ penalized the magnitude of axis-wise accelerations and steering angle, respectively. We leveraged the differential flatness property to express steering angle as a function of axis-wise velocity and acceleration terms. As can be seen from Fig.\ref{com_plot}, D-PRIEST maintained three different distributions (shown in green, red, and blue) in parallel leading to multi-modal behaviors. Intuitively, the obtained maneuvers correspond to overtaking the static obstacles (shown in black) from left to right or slowing down and shifting to another lane. In contrast, the traditional CEM was able to obtain only a single maneuver for the vehicle.

\subsection{Quantiative Results}
\subsubsection{Comparison with MPPI, Log-MPPI, TEB, DWA}
Table \eqref{Barn} summarizes the quantitative results. PRIEST reaches a 90\% success rate while the best baseline (TEB) achieves 83\%. The latter (and  DWA as well) uses graph search along with complex recovery maneuvers for an improved success rate, albeit at the expense of increased travel time. In contrast, purely optimization-based approaches MPPI and log-MPPI perform 32\% and 35\% worse than PRIEST in success rate, accompanied by marginally higher travel times as well. PRIEST shows a slightly higher mean computation time but remains fast enough for real-time applications. 

\vspace{-0.13cm}
\begin{table}[h] 
\caption{Comparisons on the BARN Dataset}\label{Barn}
\scriptsize
\centering
\begin{tabular}{|l|c|c|c|c|}
\hline
                             Method    & \cellcolor[HTML]{CBCEFB}\begin{tabular}[c]{@{}c@{}}Success\\  rate\end{tabular} & \cellcolor[HTML]{CBCEFB}\begin{tabular}[c]{@{}c@{}}Travel time\\    Mean/ Min/Max\end{tabular}  & \cellcolor[HTML]{CBCEFB}\begin{tabular}[c]{@{}c@{}}Computation  time\\ Mean/ Min/Max\end{tabular} \\ \hline
\cellcolor[HTML]{FFCCC9}DWA      & 76\%                                                                          & \begin{tabular}[c]{@{}c@{}}52.07/ 33.08/145.79\end{tabular}                                                                   & \begin{tabular}[c]{@{}c@{}}0.037/ 0.035/0.04\end{tabular}                                            \\ \hline
\cellcolor[HTML]{FFCCC9}TEB      & 83\%                                                                          & \begin{tabular}[c]{@{}c@{}}52.34/ 42.25/106.32\end{tabular}                                                                        & \begin{tabular}[c]{@{}c@{}}0.039/0.035/0.04\end{tabular}                                            \\ \hline
\cellcolor[HTML]{FFCCC9}MPPI     & 58\%                                                                            & \begin{tabular}[c]{@{}c@{}}36.66/ 31.15/99.62\end{tabular}                                                                         & \begin{tabular}[c]{@{}c@{}}0.019/0.018/0.02\end{tabular}                                            \\ \hline
\cellcolor[HTML]{FFCCC9}log-MPPI & 55\%                                                                            & \begin{tabular}[c]{@{}c@{}}36.27/30.36/58.84\end{tabular}                                                                          & \begin{tabular}[c]{@{}c@{}}0.019/0.018/0.02\end{tabular}                                            \\ \hline
\cellcolor[HTML]{FFCCC9}\textbf{Ours}     & \textbf{90\%}                                                                 & \begin{tabular}[c]{@{}c@{}}\textbf{33.59}/ 30.03/70.98\end{tabular}                                                                           & \begin{tabular}[c]{@{}c@{}}0.071/0.06/0.076\end{tabular}                                            \\ \hline
\end{tabular}
\normalsize
\vspace{-0.27cm}
\end{table}

\subsubsection{Comparison with Additional Gradient-Based and Sampling-based Optimizers}

Table \ref{grad_based} compares the performance of the PRIEST with all the baselines in 2D and 3D cluttered environments (recall Fig.\ref{traj_plot}). ROCKIT and FATROP were initialized with simple straight-line trajectories between the start and the goal that were typically not collision-free. Due to conflicting gradients from the neighboring obstacle, both these methods often failed to obtain a collision-free trajectory. Interestingly, the sampling-based approaches didn't fare particularly better as both CEM and VPSTO reported a large number of failures. We attribute the failures of VPSTO and CEM to two reasons. First, most of the sampled trajectories for both CEM and VPSTO fell into the high-cost/infeasible area, and as discussed before, this creates a pathologically difficult case for sampling-based optimizers. Second, both CEM and VPSTO  roll constraints into the cost as penalties and can be really sensitive to tuning the individual cost terms. In summary, the success-rate trend of Table \ref{grad_based} shows how both classes of both gradient and sampling-based approaches struggle in highly cluttered environments. Consequently, a strong potential solution is to use something like PRIEST that can guide trajectory sampling with convex optimization toward constraint satisfaction.

PRIEST also shows superior computation time than all the baselines. ROCKIT is based on sequential quadratic programming and scales cubically with a number of constraints. FATROP performs comparatively better as it adopts an unconstrained, penalty-based approach. The CEM run times are comparable to PRIEST. Although VPSTO numbers are high, we note that the original author implementation that we use may not have been optimized for computation speed.

\subsubsection{Combination of Gradient-Based and Sampling-based Optimizers}\label{comb_grad_sample}

\noindent A simpler alternative to PRIEST can be just to use a sampling-based optimizer to compute a good guess for the gradient-based solvers \cite{kim2022mppi}. However, such an approach will only be suitable for problems with differentiable costs. Nevertheless, we evaluate this alternative for the point-to-point benchmark of Fig.\ref{traj_plot}. We used CEM to compute an initial guess for ROCKIT and FATROP. The results are summarized in Table \ref{sto_based}. As can be seen, while the performance of both ROCKIT and FATROP improved in 2D environments, the success rate of the latter decreased substantially in the 3D variant. The main reason for this conflicting trend is that the CEM (or any initial guess generation) is unaware of the exact capabilities of the downstream gradient-based optimizer. It should also be noted that the trend of Table \eqref{sto_based} is not general and on some other problems, FATROP might outperform ROCKIT. This unreliability forms the core motivation behind PRIEST, which outperforms all ablations in Table \eqref{sto_based}. By embedding the projection optimizer within the sampling process itself (refer Alg.\ref{algo1}) and augmenting the projection residual to the cost function, we ensure that the sampling and projection complement each other.

\begin{table}[b]
\vspace{-0.35cm}
\caption{Comparing PRIEST with Gradient/Sampling-Based Optimizers }
\label{grad_based}
\centering
\scriptsize
\begin{tabular}{|c|c|c|c|}
\hline
Method & Success rate  & \begin{tabular}[c]{@{}c@{}}Computation time(Mean/Min/Max)\end{tabular} \\ \hline
\rowcolor[HTML]{8CFF8C} 
ROCKIT-2D & 46\%          & 2.57/0.6/6.2                       \\ \hline
\rowcolor[HTML]{8CFF8C} 
FATROP-2D & 64\%          & 0.63/0.07/2.87                                                           \\ \hline
\rowcolor[HTML]{8CFF8C} 
\textbf{Ours-2D}   & \textbf{95\%} & \textbf{0.043/0.038/0.064}                            
\\ \hline
\rowcolor[HTML]{8CFF8C} 
CEM-2D   & 78\%         & 0.017/0.01/0.03                                                              \\ \hline
\rowcolor[HTML]{8CFF8C} 
VPSTO-2D & 66\%         & 1.63/0.78/4.5                                                                                      \\ \hline
\rowcolor[HTML]{FFFFC7} 
ROCKIT-3D &      65\%     &      1.65/0.68/5                                                                            \\ \hline
\rowcolor[HTML]{FFFFC7} 
FATROP-3D &    81\%      &    0.088/0.034/0.23                                                                    \\ \hline
\rowcolor[HTML]{FFFFC7} 
\textbf{Ours-3D}   & \textbf{90\%} & \textbf{0.053/0.044/0.063}                                                       \\ \hline
\rowcolor[HTML]{FFFFC7} 
CEM-3D   & 74\%         & 0.028/0.026/0.033                                      \\ \hline
\rowcolor[HTML]{FFFFC7} 
VPSTO-3D & 37\%          & 3.5/0.93/3.5                                  \\ \hline
\end{tabular}
\normalsize
\vspace{-0.3cm}
\end{table}


\begin{table}[b]
\caption{Comparing PRIEST with Hybrid Gradient-Sampling Baselines. }
\scriptsize
\centering
\label{sto_based}
\begin{tabular}{|c|c|c|c|}
\hline
Method   & Success rate & \begin{tabular}[c]{@{}c@{}}Computation time (Mean/Min/Max)\end{tabular}  \\ \hline
\rowcolor[HTML]{8CFF8C} 
ROCKIT-CEM & 94\%         & 2.07/0.69/6.0                                                         \\ \hline
\rowcolor[HTML]{8CFF8C} 
FATROP-CEM & 84\%          & 0.34/0.06/0.96   
\\ \hline
\rowcolor[HTML]{8CFF8C} 
\textbf{Ours-2D}   & \textbf{95\%} & \textbf{0.043/0.038/0.064}                     
\\ \hline
\rowcolor[HTML]{FFFFC7} 
ROCKIT-CEM & 100\%          & 1.19/0.7/2.56                                             \\ \hline
\rowcolor[HTML]{FFFFC7} 
FATROP-CEM & 25\%          & 0.056/0.039/0.079                                  \\ \hline
\rowcolor[HTML]{FFFFC7} 
\textbf{Ours-3D}   & \textbf{90\%} & \textbf{0.053/0.044/0.063}                              \\ \hline                       
\end{tabular}
\normalsize
\vspace{-0.5cm}
\end{table}

\subsubsection{Benchmarking in Dynamic Environments}

Table \eqref{dynamic} presents the results obtained from the experiments in dynamic environments. By having a success rate of 83\%, our method outperforms other approaches. Furthermore, our method shows competitive efficiency with a mean travel time of 11.95 seconds. Overall, the results show the superiority of our approach in dealing with the complexities of cluttered dynamic environments, making it a promising solution for real-world applications in human-habitable environments.

\subsubsection{Scaling of D-PRIEST} 
\begin{figure}[t]
  \begin{minipage}[!t]{.47\linewidth}
   \hspace{-1.0cm}
    \includegraphics[scale=0.054]{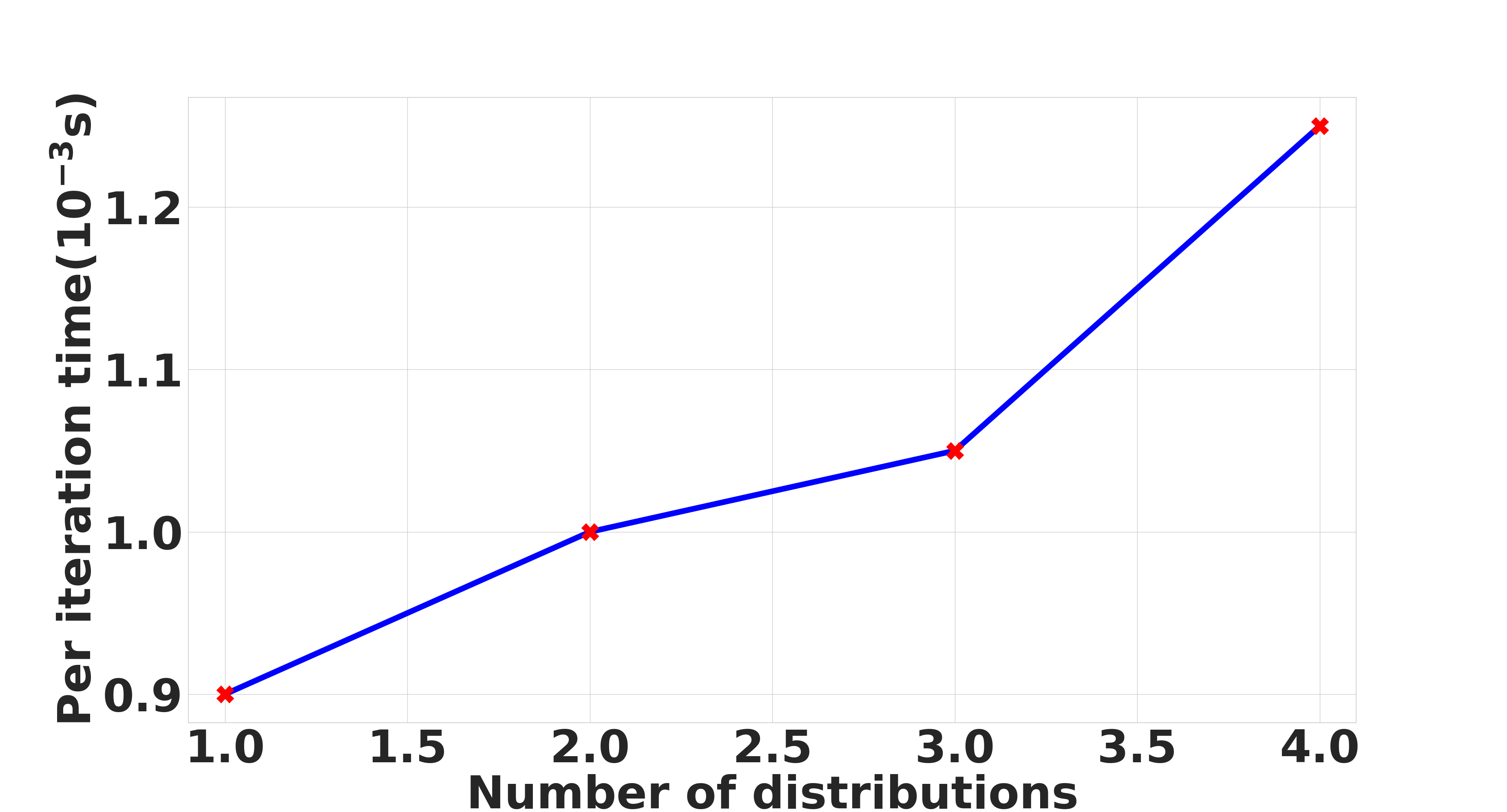}%
    \caption
      { Linear computation-time scaling of decentralized PRIEST with respect to the number of parallel distributions maintained at each iteration
        \label{comp_decenter}%
      }%
  \end{minipage}\hfill
  \begin{minipage}[!t]{.45\linewidth}  
\scriptsize
\begin{tabular}{|
>{\columncolor[HTML]{FFCCC9}}p{0.99cm}|p{0.5cm}|p{1.0cm}|p{0.1cm}|}
\hline
\cellcolor[HTML]{FFFFFF}Method & \cellcolor[HTML]{9698ED}\begin{tabular}[c]{@{}l@{}}Succ.\\ rate \end{tabular} & \cellcolor[HTML]{9698ED}\begin{tabular}[c]{@{}l@{}}Travel time \\mean\\min/max\\ \end{tabular} \\ \hline
log-MPPI                       &60\%                                 & \begin{tabular}[c]{@{}l@{}}18.80\\15.68/24.3 \end{tabular}                                                                            \\ \hline
MPPI                           & 53\%                                 &\begin{tabular}[c]{@{}l@{}} 19.38\\16.28/27.08                                        \end{tabular}                                \\ \hline
CEM                            & 46\%                                 & \begin{tabular}[c]{@{}l@{}}11.95\\9.57/14.21 
\end{tabular}\\ \hline
DWA                            & 66\%                                 &       \begin{tabular}[c]{@{}l@{}}33.4\\31.4/37.17  
\end{tabular}\\ \hline
\textbf{Ours}                           & \textbf{83\%}                        & \begin{tabular}[c]{@{}l@{}}\textbf{11.95} \\\textbf{11.43/13.39}                              \end{tabular}                                            \\ \hline
\end{tabular}
\captionof{table}{Comparisons in cluttered and dynamic environments \label{dynamic}}
  \end{minipage}
  \vspace{-0.5cm}
\end{figure}

Fig.\eqref{comp_decenter} shows the linear scaling of the per-iteration time of D-PRIEST with respect to the number of distributions. Typically, solutions are obtained in around 20 iterations. Thus, around 4 parallel distributions can be maintained under real-time constraints.

 \normalsize

\section{Conclusions and Future Work}
We presented PRIEST, an important contribution towards leveraging the benefits of both sampling-based optimizer and convex optimization. In particular, we used the latter to derive a GPU-accelerated projection optimizer that guides the trajectory sampling process. We also showed how the same projection set-up can be easily embedded within decentralized variants of sampling optimizers wherein multiple parallel distributions are maintained at each iteration. We performed a very extensive benchmarking showcasing the benefits of PRIEST over SOTA approaches in the context of autonomous navigation in unknown environments. Our future efforts are focused on extending PRIEST to high-dimensional manipulation problems.

\section{Appendix}\label{append}

\noindent \textbf{Reformulating constraints:} We reformulate collision avoidance inequality constraints \eqref{coll_multiagent} as follows: 

\vspace{-0.3cm}
\small
\begin{align}
\hspace{-0.17cm}\textbf{f}{_{o,j}}\hspace{-0.11cm} = \hspace{-0.15cm}
\left \{\hspace{-0.05cm} \begin{array}{lcr}\hspace{-0.21cm}
x\hspace{-0.01cm}(t)\hspace{-0.09cm}-\hspace{-0.09cm}x_{o,j}\hspace{-0.02cm}(t)\hspace{-0.1cm}-\hspace{-0.08cm} ad_{o,j}\hspace{-0.02cm}(t)\hspace{-0.02cm}\cos{\alpha_{o,j}}\hspace{-0.02cm}(t)\hspace{-0.01cm}\sin{\beta_{o,j}}\hspace{-0.02cm}(t) \\
\hspace{-0.21cm}y\hspace{-0.01cm}(t)\hspace{-0.08cm}-\hspace{-0.08cm}y_{o,j}\hspace{-0.02cm}(t) 
\hspace{-0.08cm}-\hspace{-0.08cm}ad_{o,j}\hspace{-0.02cm}(t)\hspace{-0.01cm}\sin{\alpha_{o,j}\hspace{-0.02cm}(t)}\sin{\beta_{o,j}}\hspace{-0.02cm}(t)\\
\hspace{-0.21cm}z(t)\hspace{-0.08cm}  -z_{o,j}(t)\hspace{-0.08cm} -bd_{o,j}(t)\cos{\beta_{o,j}(t)}
\end{array} \hspace{-0.30cm} \right \}\hspace{-0.12cm},\hspace{-0.06cm}d_{o,j}\hspace{-0.03cm}(t)\hspace{-0.13cm} \geq \hspace{-0.09cm}1 
\label{collision_avoidance_proposed} 
\end{align}
\normalsize

\noindent where $d_{o,j}(t),\alpha_{o,j}(t)$ and $\beta_{o,j}(t)$ are the line-of-sight~distance and angles between the robot center and $j^{th}$ obstacle (see \cite{rastgar2021gpu,rastgar2020novel}). 

Similarly, the inequality constraints \eqref{acc_constraint}, which show the maximum velocity and acceleration, can be reformulated as: 

\vspace{-0.3cm}
\small
\begin{subequations} 
\begin{align}
\hspace{-0.3cm}\textbf{f}_{v} \hspace{-0.09cm}=  \hspace{-0.13cm}\left \{ \hspace{-0.20cm} \begin{array}{lcr}
 \dot{x}(t) \hspace{-0.08cm}- \hspace{-0.04cm} d_{v}(t)v_{max}\cos{\alpha_{v}(t)}\sin{\beta_{v}(t)}  \\
 \dot{y}(t) \hspace{-0.08cm}- d_{v}(t)v_{max}\sin{\alpha_{v}(t)}\sin{\beta_{v}(t)} \\
 \dot{z}(t)\hspace{-0.07cm} -d_{v}(t)v_{max}\cos{\beta_{v}(t)}
\end{array} \hspace{-0.27cm} \right \} \hspace{-0.07cm}
,0\leq d_{v}(t) \hspace{-0.05cm}\leq \hspace{-0.05cm}1 \hspace{-0.02cm},  \hspace{-0.02cm} \label{fv} \\
\hspace{-0.20cm}\textbf{f}_{a}\hspace{-0.08cm} = \hspace{-0.13cm}\left \{ \hspace{-0.20cm}\begin{array}{lcr}
\ddot{x}(t)\hspace{-0.07cm} - d_{a}(t)a_{max}\cos{ \alpha_{a}(t)}\sin{\beta_{a}(t)} \\
\ddot{y}(t) \hspace{-0.07cm}- d_{a}(t)a_{max}\sin{\alpha_{a}(t)}\sin{\beta_{a}(t)}\\
\ddot{z}(t)\hspace{-0.07cm} - d_{a}(t)a_{max}\cos{\beta_{a}(t)}
\end{array} \hspace{-0.27cm} \right \}\hspace{-0.07cm}, 0\leq d_{a}(t)\hspace{-0.05cm} \leq 1 , \hspace{-0.05cm} \label{fa}
\end{align}
\end{subequations}
 \normalsize

\noindent Variables $d_{o,j}(t),\alpha_{o,j}(t),\beta_{o,j}(t),d_{v}(t),d_{a}(t),\alpha_{v}(t),\alpha_{a}(t)$, $\beta_{a}(t)$ and $\beta_{v}(t)$ are additional variables which will be computed along the projection part.

\noindent \textbf{Reformulated Problem:} Now, leveraging the above reformulations, we can re-phrase the projection problem \eqref{proj_cost}-\eqref{reform_bound_p} for the $i^{th}$ sample (see Alg.\ref{algo1}) as:

\vspace{-0.5cm}
\small
\begin{subequations}
\begin{align} \overline{\boldsymbol{{\xi}}}_{i} = \arg\min_{\overline{\boldsymbol{\xi}}_{i}} \frac{1}{2}\Vert \overline{\boldsymbol{\xi}}_{i}-\boldsymbol{\xi}_i\Vert_2^2 \label{cost_modify} \\
    \textbf{A}\overline{\boldsymbol{\xi}}_{i} = \textbf{b}_{eq}, \label{eq_modify_our}\\
   \Tilde{\textbf{F}}\overline{\boldsymbol{{\xi}}}_{i} = 
   \Tilde{\textbf{e}}(\boldsymbol{\alpha}_{i}, \boldsymbol{\beta}_{i}, \textbf{d}_{i})\label{reform_bound},\\
 \textbf{d}_{min} \leq \textbf{d}_{i}\leq \textbf{d}_{max}, \label{ine_1} \\
 \textbf{G}\overline{\boldsymbol{\xi}}_{i} \leq \boldsymbol{\tau} \label{bound_con}
\end{align}
\end{subequations}
\normalsize

\noindent where $\boldsymbol{\alpha}_{i}, \boldsymbol{\beta}_{i}$ and $\textbf{d}_{i}$ are the representation of \hspace{-0.4cm}$\begin{bmatrix}
      \boldsymbol{\alpha}_{o,i}^{T} &\hspace{-0.2cm}\boldsymbol{\alpha}_{v,i}^{T}&\hspace{-0.2cm}
      \boldsymbol{\alpha}_{a,i}^{T}, 
  \end{bmatrix}^{T}\hspace{-0.2cm}, \begin{bmatrix}
      \boldsymbol{\beta}_{o,i}^{T} &\hspace{-0.2cm}\boldsymbol{\beta}_{v,i}^{T} &
      \boldsymbol{\beta}_{a,i}^{T} \hspace{-0.2cm}
  \end{bmatrix}^{T}\hspace{-0.2cm}, \text{and}
     \begin{bmatrix}
     \textbf{d}^{T}_{o,i}&\hspace{-0.2cm}
     \textbf{d}^{T}_{v,i} &\hspace{-0.2cm} \textbf{d}^{T}_{a,i}
  \end{bmatrix}^{T}$ respectively. The constant vector $\boldsymbol{\tau}$ is formed by stacking the $s_{min}$ and $s_{max}$ in appropriate form. The matrix $\textbf{G}$ is formed by stacking $ -\bold{P}$ and $\bold{P}$ vertically. Similarly, $\textbf{d}_{min}$, $\textbf{d}_{max}$ are formed by stacking the lower ($[1, 0, 0]$), and upper bounds ($[\infty, 1, 1]$) of $\textbf{d}_{o,i}, \textbf{d}_{v,i}, \textbf{d}_{a,i}$. Also, $ \Tilde{\textbf{F}}$, and $\textbf{e}$ are formed as

\vspace{-0.3cm}
  \small
  \begin{align}
      \tilde{\textbf{F}}\hspace{-0.10cm} =\hspace{-0.15cm} \begin{bmatrix}
        \begin{bmatrix}
        \textbf{F}_{o} \\
        \dot{\bold{P}} \\
        \ddot{\bold{P}}
    \end{bmatrix} \hspace{-0.25cm}&\hspace{-0.25cm} \textbf{0} \hspace{-0.25cm}&\hspace{-0.25cm} \textbf{0} \\ \textbf{0}\hspace{-0.25cm} & \hspace{-0.25cm}\hspace{-0.15cm}\begin{bmatrix}
        \textbf{F}_{o} \\
        \dot{\bold{P}} \\
        \ddot{\bold{P}}
    \end{bmatrix} \hspace{-0.25cm}& \hspace{-0.25cm} \textbf{0}\\
         \textbf{0}\hspace{-0.25cm}& \hspace{-0.25cm}\textbf{0} \hspace{-0.25cm}&\hspace{-0.25cm} \begin{bmatrix}
        \textbf{F}_{o} \\
        \dot{\bold{P}} \\
        \ddot{\bold{P}}
    \end{bmatrix}
    \end{bmatrix}\hspace{-0.15cm}, \hspace{-0.05cm}
    \Tilde{\textbf{e}} \hspace{-0.1cm}= \hspace{-0.15cm}\begin{bmatrix}
        \textbf{x}_{o} +
      a\textbf{d}_{o,i}\cos{ \boldsymbol{\alpha}_{o,i}} \sin{ \boldsymbol{\beta}_{o,i}} \\ \textbf{d}_{v,i}v_{max}\cos{\boldsymbol{\alpha}_{v,i}}\sin{\boldsymbol{\beta}_{v,i}} \\
\textbf{d}_{a,i}a_{max}\cos{\boldsymbol{\alpha}_{a,i}}\sin{\boldsymbol{\beta}_{a,i}}\\
        \textbf{y}_{o} +
      a\textbf{d}_{o,i}\sin{ \boldsymbol{\alpha}_{o,i}} \sin{ \boldsymbol{\beta}_{o,i}} \\
\textbf{d}_{v,i}v_{max}\sin{\boldsymbol{\alpha}_{v,i}}\sin{\boldsymbol{\beta}_{v,i}}\\
\textbf{d}_{a,i}a_{max}\sin{\boldsymbol{\alpha}_{a,i}}\sin{\boldsymbol{\beta}_{a,i}}\\
      \textbf{z}_{o} +
      b\hspace{0.1cm} \textbf{d}_{o,i}\cos{ \boldsymbol{\beta}_{o,i}} \\
\textbf{d}_{v,i}v_{max}\cos{\boldsymbol{\beta}_{v,i}}\\
\textbf{d}_{a,i}a_{max}\cos{\boldsymbol{\beta}_{a,i}}
    \end{bmatrix}\hspace{-0.15cm},\label{f_e}
  \end{align} 
  \normalsize

\noindent where $\textbf{F}_{o}$ is formed by stacking as many times as the number of obstacles. Also, $\textbf{x}_{o}, \textbf{y}_{o}, \textbf{z}_{o}$ is obtained by stacking $x_{o,j}(t),y_{o,j}(t), z_{o,j}(t)$ at different time stamps and for all obstacles. 

\noindent\textbf{Solution process}
We utilize the augmented Lagrangian method to relax the equality and affine constraints \eqref{reform_bound}-\eqref{bound_con} as $l_{2}$ penalties. Consequently, the projection cost can be rephrased as follows:

\vspace{-0.3cm}
\small
\begin{align}
\hspace{-0.20cm}\mathcal{L}\hspace{-0.1cm}=\hspace{-0.1cm}\frac{1}{2}\hspace{-0.1cm}\left\Vert \overline{\boldsymbol{\xi}}_{i}\hspace{-0.1cm} - \hspace{-0.1cm}\boldsymbol{\xi}_{i}\right\Vert ^{2}_{2}\hspace{-0.1cm}- \hspace{-0.13cm}\langle\boldsymbol{\lambda}_{i}, \overline{\boldsymbol{\xi}}_{i}\rangle \hspace{-0.1cm}
 +\hspace{-0.1cm}\frac{\rho}{2}\left\Vert \Tilde{\textbf{F}}\overline{\boldsymbol{\xi}}_{i}  
 \hspace{-0.15cm}-\Tilde{\textbf{e}} \right \Vert_2^2 
 \hspace{-0.1cm}+\hspace{-0.1cm}\frac{\rho}{2}\left\Vert \textbf{G}\overline{\boldsymbol{\xi}}_{i}  
 \hspace{-0.15cm}-\boldsymbol{\tau}+\hspace{-0.08cm}\textbf{s}_{i} \right \Vert_2^2 , \nonumber \\
 =\hspace{-0.1cm} \frac{1}{2}\left\Vert \overline{\boldsymbol{\xi}}_{i} - \boldsymbol{\xi}_{i}\right\Vert ^{2}_{2}\hspace{-0.1cm}- \hspace{-0.1cm}\langle\boldsymbol{\lambda}_{i}, \overline{\boldsymbol{\xi}}_{i}\rangle \hspace{-0.1cm}
 +\hspace{-0.1cm}\frac{\rho}{2}\left\Vert \textbf{F}\overline{\boldsymbol{\xi}}_{i}  
 \hspace{-0.15cm}-\textbf{e} \right \Vert_2^2 
 \label{lagrange} 
 \end{align}
\normalsize

\noindent where,
$\hspace{-0.05cm}\textbf{F} = \hspace{-0.1cm}\begin{bmatrix}
    \Tilde{\textbf{F}} \\ \textbf{G}
\end{bmatrix}\hspace{-0.1cm},
\textbf{e} =\hspace{-0.1cm}\begin{bmatrix}
   \Tilde{\textbf{e}} \\ \boldsymbol{\tau}\hspace{-0.1cm}-\textbf{s}_{i}
\end{bmatrix}$. Also, $\boldsymbol{\lambda}_{i}$, $\rho$ and $\textbf{s}_{i}$ are Lagrange multiplier, scalar constant and slack variable. We minimize \eqref{lagrange} subject to \eqref{eq_modify_our} using AM, which is reduced to the following steps. 

\vspace{-0.3cm}
\small
\begin{subequations}
\begin{align}
    ^{k+1}\boldsymbol{\alpha}_{i} = \text{arg} \min_{\boldsymbol{\alpha}_{i}}\mathcal{L}(^{k}\overline{\boldsymbol{\xi}}_{i}, \boldsymbol{\alpha}_{i}, \hspace{0.1cm}^{k}\boldsymbol{\beta}_{i}, \hspace{0.1cm}^{k}\textbf{d}_{i}, \hspace{0.1cm}^{k}\boldsymbol{\lambda}_{i},
    \hspace{0.1cm}^{k}\textbf{s}_{i} ) \label{comp_alpha} \\
    ^{k+1}\boldsymbol{\beta}_{i} = \text{arg} \min_{\boldsymbol{\beta}_{i}}\mathcal{L}(^{k}\overline{\boldsymbol{\xi}}_{i}, \hspace{0.1cm}^{k+1}\boldsymbol{\alpha}_{i}, \boldsymbol{\beta}_{i}, \hspace{0.1cm}^{k}\textbf{d}_{i}, \hspace{0.1cm}^{k}\boldsymbol{\lambda}_{i},
    \hspace{0.1cm}^{k}\textbf{s}_{i} ) \label{comp_beta}\\
    ^{k+1}\textbf{d}_{i} = \text{arg} \min_{\textbf{d}_{i}}\mathcal{L}(\hspace{0.1cm}^{k}\overline{\boldsymbol{\xi}}_{i}, \hspace{0.1cm}^{k+1}\boldsymbol{\alpha}_{i}, \hspace{0.1cm}^{k+1}\boldsymbol{\beta}_{i},\textbf{d}_{i}, \hspace{0.1cm}^{k}\boldsymbol{\lambda}_{i},
    \hspace{0.1cm}^{k}\textbf{s}_{i} )\label{comp_d}\\
     ^{k+1}\textbf{s}_{i} = \max (\textbf{0}, -\textbf{G}\hspace{0.1cm}^{k}\overline{\boldsymbol{\xi}}_{i}+ \boldsymbol{\tau})\\
    \hspace{0.1cm}^{k+1}\boldsymbol{\lambda}_{i} = ^{k}\boldsymbol{\lambda}_{i}
- \rho \textbf{F}^{T}(\textbf{F}\hspace{0.1cm} ^{k}\overline{\boldsymbol{\xi}}_{i} 
 -^{k}\Tilde{\textbf{e}})\label{lag} \\
     ^{k+1}\textbf{e}=
     \begin{bmatrix}
        \Tilde{\textbf{e}}(\hspace{0.1cm}^{k+1}\boldsymbol{\alpha}_{i},^{k+1}\boldsymbol{\beta}_{i}, ^{k+1}\textbf{d}_{i}) \label{obtain_e}\\
    \boldsymbol{\tau} -^{k+1}\textbf{s}_{i}\end{bmatrix}\\
  ^{k+1}\overline{\boldsymbol{\xi}}_{i} = \text{arg} \min_{\overline{\boldsymbol{\xi}}_{i}}\mathcal{L}({\boldsymbol{\xi}}_{i}, {^{k+1}}\textbf{e}_i,  ^{k+1}\boldsymbol{\lambda}_{i},
) \label{comp_zeta}
\end{align}
\end{subequations}
\normalsize

\vspace{-0.2cm}

\noindent For each AM step, we only optimize one group of variables while others are held fixed. Note that stacking of right-hand sides of \eqref{obtain_e} and \eqref{lag} provide the function $\textbf{h}$ presented in \eqref{fixed_point_1}. The steps \eqref{comp_alpha}-\eqref{comp_d} have closed form solutions in terms of ${^k}\overline{\boldsymbol{\xi}}_i$ \cite{rastgar2020novel,rastgar2021gpu, masnavi2022visibility}. Also, \eqref{comp_zeta} is a representation of \eqref{proj_cost}-\eqref{reform_bound_p}.

\noindent \begin{remak} \label{matrix_scaling}
The matrix $\textbf{F}$ and vector $\textbf{e}$ (see \eqref{f_e}) have the dimension of $3(n_{o}+2)n_{p} \times 3n_{v}$ and $3(n_{0}+2)n_{p}$ and their complexity grow linearly with the number of obstacles $n_o$ and the planning horizon $n_p$. 
\end{remak}



\bibliography{iros_ral_2022_ref}
\footnotesize{
\bibliographystyle{IEEEtran}
}

\end{document}